\documentclass[10pt,a4paper,twoside]{article}
\usepackage{latexsym,amssymb}
\usepackage{graphicx,inputenc}
\usepackage{url}

\begin{document}

\begin{center}{\Large\bf
Enhancing Apparent Personality Trait Analysis with Cross-Modal Embeddings
}\end{center}

\begin{center}{\large\bf\noindent
Ádám Fodor, Rachid R. Saboundji, András Lőrincz
}\\[2mm]
Department of Artificial Intelligence, Faculty of Informatics, Eötvös Loránd University, Budapest, Hungary
\\[1mm]\texttt{
foauaai@inf.elte.hu, sxdj3m@inf.elte.hu, lorincz@inf.elte.hu
}\end{center}

\vspace*{7mm}

\begin{abstract}
Automatic personality trait assessment is essential for \mbox{high-quality} \mbox{human-machine} interactions.
Systems capable of human behavior analysis could be used for self-driving cars, medical research, and surveillance, among many others. We present a multimodal deep neural network with a Siamese extension for apparent personality trait prediction trained on short video recordings and exploiting modality invariant embeddings. Acoustic, visual, and textual information are utilized to reach high-performance solutions in this task. Due to the highly centralized target distribution of the analyzed dataset, the changes in the third digit are relevant. Our proposed method addresses the challenge of under-represented extreme values, achieves \mbox{$0.0033$ MAE} average improvement, and shows a clear advantage over the baseline multimodal DNN without the introduced module.
\end{abstract}

\section{Introduction}
Prediction of personality traits is an important task since it is useful for predicting decision-making patterns of people with stable personality traits in diverse situations and detecting changes due to, e.g., stress, drinking, drugs, and so on. One of the most studied model to describe personality is the Big Five personality traits \cite{digman1990personality}. The theory identifies five factors: EXTraversion, NEUroticism, AGReeableness, CONscientiousness, and OPEnness. Each personality trait represents a range bounded by two extremes, e.g., for extraversion, the two ends are extreme extraversion and extreme introversion.

Audio-visual personality trait prediction has become of high-interest \cite{crnet2020} due to high-quality databases released in the ChaLearn challenges, i.e., in First Impressions V1 and V2 \cite{firstimpression2018}. In this study, we used the extended and revised dataset (V2). The dataset contains 10,000 video clips extracted from more than 3,000 different YouTube high-definition videos of people mostly facing and speaking to a camera.

Although multimodal systems offer advantages compared to monomodal systems, they raise several challenges as well. For example, one faces the problems of selecting from the modalities to be included into multimodal systems, deriving the architecture to fuse them, and attenuating errors from noisy, missing, or underrepresented data. One specific characteristic of the First Impression V2 database is its unbalanced data distribution with fewer extreme samples. However, these examples have much more significance and have priority in several use cases, including medication.

Multimodal fusion approaches often hardly consider complex intra- and inter-modal dependencies and lack robustness in case of noisy or missing modalities \cite{zadeh2017tensor}. Due to these challenges, an increasing number of studies were conducted to transfer knowledge across domains or modalities \cite{kang2015learning, wang2017adversarial}. Embedding methods have been proven useful for overcoming the before-mentioned inter-dependencies.  It has been found that similarity and correlation of semantic information retrieved from real data can be represented using deep metric learning in an \mbox{embedded feature space \cite{hoffer2015deep, han2019emobed}}.

\newpage

Our contributions are listed below:
\begin{enumerate}
\item We propose a general-purpose learning framework to extract \mbox{modality-invariant embeddings} from multiple information sources with a Siamese network, emphasizing extreme examples and implicitly improving the multimodal fusion process.

\item We extended the Multi-Similarity loss \cite{wang2019multi} to handle multiple apparent personality trait class labels simultaneously, besides using various input modalities. The problem with non-extreme examples that one or more modalities contain inadequate information to aid the deep embedding process. To overcome this issue, we modified the sample selection of the so called ``online hard example mining procedure'' of the triplet loss evaluation and put the emphasis onto the extreme samples to be detailed in the paper.

\item Although samples having lower or higher personality trait values are less frequent in the database, high quality prediction of their values is desired in various situations. We show that cross-modal embedding enhances the prediction of the Big Five personality traits in the extreme cases.
\end{enumerate}

The paper is organized as follows. Section~\ref{s:relwork} reviews the related works. The preprocessing steps, baseline, and the proposed method are detailed in Section~\ref{s:methodology}. The experimental setup, dataset introduction, and the implementation details are described in Section~\ref{s:exp_results}. Our results, together with the discussion are presented in Section~\ref{s:results}. We conclude in Section~\ref{s:conc}.  

\section{Related works}\label{s:relwork}

Multimodal information has been widely used in various domains ranging from semantic indexing, multimedia event detection to video situation understanding, among many others. To merge such sources of information, fusion strategies have been derived to harness complementary information from single modalities. Such strategies are classified into three categories, model-level fusion, feature-level fusion, and decision level fusion \cite{zeng2008survey}.

Human behavior monitoring and evaluation rely heavily on multimodal information fusion. Busso et al. \cite{busso2004analysis} paired facial expressions with audio information yielding better prediction for emotion recognition.  Wimmer, et al. \cite{bjorn2008low} studied feature-level fusion of low-level audio and video description. Contextual long-range information was later leveraged by the introduction of BLSTMs by W{\"o}llmer et al \cite{bjorn2010context}.  In contrast, with the emergence of deep learning, more sophisticated methods were adopted, e.g., by Ngiam et al. \cite{andrew2011multimodal}, who suggested a bi-modal deep auto-encoder to extract shared representations from the input modalities. However, these approaches hardly consider complex intra- and inter-modal interactions and lack robustness in case of noisy or missing modalities \cite{lpmorency2017tensor,andrew2011multimodal}.

Embedding methods have been proven useful for integrating such inter-dependencies. Han, J. et al. \cite{bjorn2018emotion} used triplet loss to distill discriminative representations in the speech modality. Tsai et al. \cite{lpmorency2018learning} proposed a model that factorizes learned multimodal representations into two sets of independent generative and discriminative factors. Recently, Han et al. \cite{han2019emobed} introduced a novel learning framework to leverage information from auxiliary modalities for emotion recognition, using triplet loss to produce modality-invariant emotion embedding in a latent space.

There are recent surveys on personality trait detection that can orient the interested reader \cite{hugo2018explaining,junior2019first}. Here, we mention the works related to the challenges called ChaLearn: First Impression Challenge. Kaya et al. \cite{kaya2017multi}, the winner of the ChaLearn: First Impressions V1 competition, used visual, audio and scene features in their system trained end-to-end. Kampman et al. \cite{omno2018investigating}  performed an ablation study by combining audio, video, and text information in a tri-modal stacked CNN architecture. More recently, Zhang et al. \cite{zhang2019persemon} studied the feasibility of merging apparent personality and emotion estimations within a single deep neural network in a multi-task learning framework. An apparent problem with this approach is that the standard deviation of the estimations when trained on the  ChaLearn First impression dataset is much narrower than that of the original data. The phenomenon is called the ``regression-to-the-mean problem'' where extreme values prediction becomes severely constrained. Li et al. \cite{crnet2020} considered this issue and proposed a classification-regression model in which the final regression is guided by the learned classification features and introduced a new objective function (called Bell loss) to ease the aforementioned problem. 

\begin{figure}[!ht]
\includegraphics[width=\textwidth]{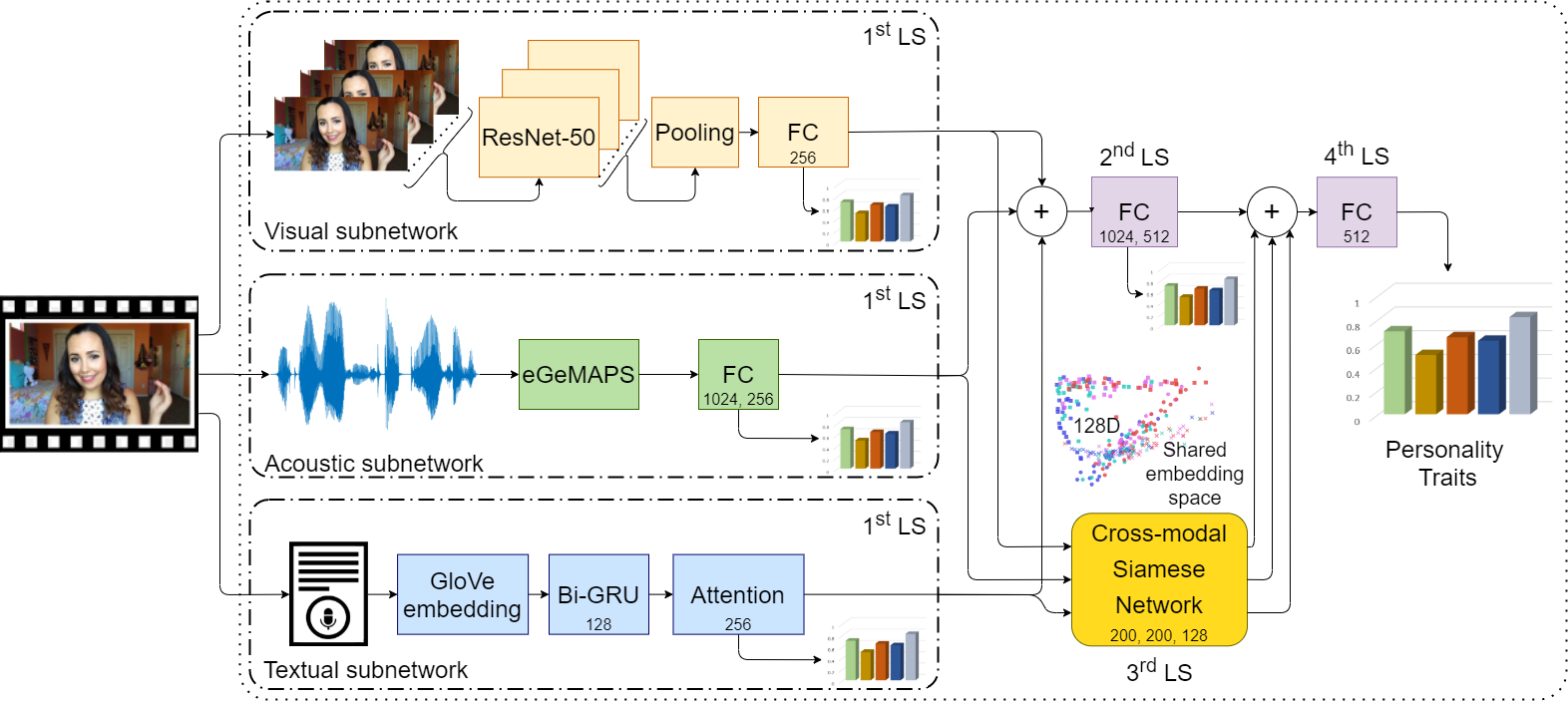}
\caption{Pipeline of the proposed method for enhanced Big Five personality trait prediction. Visual, acoustic, and textual information are processed with modality-specific subnetworks. The hidden representations are projected into a shared embedding space with a Siamese network to exploit mutual information of different information sources implicitly. The shared embedding space of the 128D auxiliary vectors is illustrated by colored circles in 2D. The extracted multimodal hidden representations and the cross-modal embeddings are fused before the final Big Five prediction. The training procedure consists of multiple learning stages (LS). FC: fully-connected, Bi-GRU: bidirectional gated recurrent unit, $\oplus$: concatenation operator. The numbers within blocks indicate the number of hidden units used. Multiple values imply stacked layers.}\label{fig:proposed}
\end{figure}

\section{Methodology}
\label{s:methodology}

In this work, we propose a multimodal deep neural network that combines features from visual, acoustic, and textual clues to predict apparent Big-Five personality traits using short video clips from the ChaLearn challenges. The pipeline is depicted in Figure~\ref{fig:proposed}.

In the case of audio signals we use standard acoustic features that can be generated by OpenSMILE \cite{eyben2010opensmile}, see later. For the visual feed, most of the frames contain redundant information and we subsample the frames. Since annotated transcripts are noisy, we adopt non-contextual word-level representation for capturing the semantic meaning.

We aim to create a shared coordinate space, transforming the audio, video, and text descriptors into a semantically relevant form using a Siamese network. The triplet-based loss functions are designed to encourage positive examples as close as possible to the so called anchor sample, and negative examples to be separated from each other over a given threshold. Embedded vector and auxiliary vector are interchangeably used for the outputs of the Siamese network. Higher precision estimation of the extremes is one of our goals and we expect that multi-modal data enrichment is advantageous in each trait. We use a DNN that combines tri-modal features along with the embedded vectors to predict apparent personality traits from the short video clips.

\subsection{Data preprocessing}

\paragraph{Audio Features}
For acoustic features, we used a de-facto standard preset called extended Geneva Minimalistic Acoustic Parameter Set (eGeMAPS) \cite{eyben2015geneva}. This feature set contains the F0 semitone, loudness, spectral flux, MFCC, jitter, shimmer, F1, F2, F3, alpha ratio, Hammarberg index, slope V0 features. Furthermore, many statistical functions are applied to these low-level descriptors considering voiced and unvoiced regions, resulting in 88 features for every sample.
The audio signals were extracted from the videos using FFmpeg with 44100 sampling frequency. Then, the eGeMAPS were generated through \mbox{OpenSMILE}. Min-max normalization was applied as a preprocessing step to rescale variables into the range [0, 1].

\paragraph{Visual Features}
We subsampled the video: only 6 frames are selected to reduce the overall complexity and redundancy of successive, similar frames. The choice of 6 is arbitrary and it does not affect the outcome significantly. Pixel values fit into the range of [0, 255]. Images are resized to $140\times248$ pixels to preserve the original aspect ratio, then the same random $128\times128$ pixels spatial crop was applied on all frames of a sample. We employed the same augmentation techniques on every frame of a single clip (with 0.5 probability) during training to preserve the relative similarity between video frames. Data augmentation consists of random flip, random hue ($\pm0.15$), brightness ($\pm0.2$), saturation (between 0.8 and 1.2) and contrast (between 0.8 and 1.2). The augmentations on hue and brightness are additive, while the saturation and contrast are multiplicative. During test and validation time, center crop was applied.  Finally, the frames are scaled between [-1; 1].

\paragraph{Textual Features}
GloVe uses unsupervised learning to obtain non-contextual vector representation of words. This vector is meant to encode semantic information, such that similar words (e.g., synonyms) have similar embedding vectors. We used pre-trained embeddings (Wikipedia 2014 and Gigaword 5), which captures the overall meaning of a sentence in a relatively lesser amount of memory, and faster than contextual models (like BERT) do.
The transcripts are tokenized with SpaCy. All special characters, digits, URLs, and emails are filtered. Every token is converted to its corresponding GloVe vector before feeding it to the textual subnetwork.

\subsection{Multimodal information fusion}\label{sec:subnet_losses}

Visual, acoustic, and textual high-level attributes are combined via a model-fusion approach. Being a regression task, in the first learning stage, the modality-specific subnetworks are trained separately, using ground truth annotations. Hence, they are used as feature extractors, and the parameters of the networks are frozen during further training. In the second learning stage, the tri-modal feature vectors are concatenated and used as the input of a fully-connected network.

\paragraph{Acoustic subnetwork}
The 88-dimensional acoustic feature vector $x_A\in\rm I\!R^N$ is the input of the audio subnetwork $f_A: \rm I\!R^N \rightarrow \rm I\!R^Q$, which is, a fully-connected shallow network with two hidden layers.

\paragraph{Visual subnetwork}
Using the video samples $x_V\in\rm I\!R^{F\times H\times W\times C}$, where $F$ is the number of frames, $H$ and $W$ are the height and width spatial dimensions, $C$ is the number of channels, a feature extractor $f_V: \rm I\!R^{F\times H\times W\times C} \rightarrow \rm I\!R^Q$ is trained. We chose ResNet-50 for our visual backbone. For every frame, a 2048-dimensional feature vector is extracted. Average pooling was applied to the time dimension, followed by a fully-connected layer. 

\paragraph{Textual subnetwork}
The textual subnetwork input is $x_T\in\rm I\!R^{K\times G}$, where $K$ is the maximum sequence length, $G$ is the dimension of GloVe embeddings. A bidirectional gated recurrent unit (Bi-GRU) with attention mechanism is trained $f_T: \rm I\!R^{K\times G} \rightarrow \rm I\!R^Q$ as a feature extractor. 

The $x_A$ audio feature vector, the corresponding $x_V$ RGB frames and $x_T$ GloVe vectors are fed into their modality-specific subnetworks, producing $h_A$, $h_V$, $h_T\in \rm I\!R^Q$ hidden representations, respectively:

\begin{equation}
h_A = f_A(x_A), \; h_V = f_V(x_V), \; h_T = f_T(x_T)
\end{equation}

Let us define $p : \rm I\!R^{(\cdot)} \rightarrow \rm I\!R^5$, which is a linear mapping function, that estimate the five personality attributes from a given hidden representation. For monomodal subnetworks the process can be formalized as follows:
\begin{equation}
\hat{y} = p(h_A), \; \hat{y} = p(h_V), \; \hat{y} = p(h_T)
\end{equation}

The network parameters are optimized with Bell loss, following the work of \cite{crnet2020}. The shape of the loss function is like an inverted bell and applied to address the regression-to-the-mean problem \cite{wang2018}, which is particularly problematic in our case, where the ground truth scores follow a Gaussian distribution closely.
The Bell loss is defined as:

\begin{equation}\label{eq:bell}
\mathcal{L}_{bell} = \frac{1}{5n} \sum_{i=1}^{n} \sum_{j=1}^{5} \gamma \Big(1-e^{-\frac{(y_{ij}-\hat{y}_{ij})^2}{2\sigma^2}}\Big),
\end{equation}

\noindent
where $n$ is the number of samples, $y_{ij}$ and $\hat{y}_{ij}$ are the ground truth and prediction of $i$th sample of $j$th trait, respectively, $\sigma$ is the derivation parameter, and $\gamma$ is a scale parameter. The $\sigma$ controls the amplitude of variation, and $\gamma$ makes the loss function consistent with other used loss functions, such as the classical Mean Absolute Error (MAE) and Mean Squared Error (MSE).

\begin{equation}\label{eq:mae_mse}
\mathcal{L}_{mae} = \frac{1}{5n} \sum_{i=1}^{n} \sum_{j=1}^{5} \vert y_{ij}-\hat{y}_{ij} \vert, \;
\mathcal{L}_{mse} = \frac{1}{5n} \sum_{i=1}^{n} \sum_{j=1}^{5} \big( y_{ij}-\hat{y}_{ij} \big)^2
\end{equation}

\noindent

As empirical results showed in \cite{crnet2020}, the Bell loss has difficulties at the beginning of the optimization and shines at later optimization stages. For avoiding the issue, the sum of $\mathcal{L}_{mae}$ and $\mathcal{L}_{mse}$ guide the stochastic gradient descent algorithm in the earlier stages by producing a higher gradient. We trained the modality-specific subnetworks with $\mathcal{L}$, which is the sum of $\mathcal{L}_{mae}$, $\mathcal{L}_{mse}$ and $\mathcal{L}_{bell}$ loss functions introduced in Equation~\ref{eq:bell} and \ref{eq:mae_mse}.

\paragraph{Baseline multimodal network}
In the second learning stage, the parameters of the $f_A$ acoustic, $f_V$ visual and $f_T$ textual subnetworks are not updated. To leverage the supplementary information of multiple modalities we concatenated the $h_A$, $h_V$ and $h_T$ hidden representations and performed model-level fusion. $M_1 : \rm I\!R^{3Q} \rightarrow \rm I\!R^O$ fully-connected shallow network and $p$ is applied to get the personality trait prediction. Formally defined as:

\begin{equation}
\hat{y} = p\Big(M_1\big(h_A\oplus h_V\oplus h_T\big)\Big),
\end{equation}

\noindent
where $\oplus$ is the concatenation operator.

\subsection{Cross-modal deep metric learning}\label{sec:siamese_loss}
In the following paragraphs, we describe the metric learning framework. We can leverage complementary information from different modalities efficiently using a Siamese network.
Using the cross-modal embedding, we make the proposed model more robust to noise, so a more accurate prediction can be achieved. 
We aim to train a cross-modal Siamese network $S: \rm I\!R^Q \rightarrow \rm I\!R^E$ on the hidden representations of modality-specific nets, which project the multimodal descriptors into a shared coordinate space $\rm I\!R^E$. 

\begin{equation}
    e_A = S(h_A), \; e_V = S(h_V), \; e_T = S(h_T)
\end{equation}

\noindent
where $S$ is a Siamese network, $e_A$, $e_V$ and $e_T$ are the projected $E$-dimensional embeddings of $h_A$, $h_V$ and $h_T$ hidden representations, respectively. 

We aim to create a common cross-modal embedding space by transforming tri-modal descriptors in a semantically relevant way. For training the Siamese network, we choose the current \mbox{state-of-the-art}, triplet-base multi-similarity (MS) loss function \cite{wang2019multi}, which requires an anchor, a positive and a negative example to form positive and negative pairs within a mini-batch. It can jointly measure the self-similarity and relative similarities of a pair, which allows it to collect informative pairs by implementing iterative pair mining and weighting.
Deep metric learning requires class labels for training, and MS loss is proposed and tested for only one modality, the single RGB texture.

\paragraph{Triplet generation}
Using inputs and the corresponding class labels, we can form triplets $\{e, e^+, e^-\}$. Examples from the same class $\{e, e^+\}$ are determined as positive pairs $\in\mathcal{P}$, as well as samples belonging to different classes $\{e, e^-\}$ are the negative pairs $\in\mathcal{N}$. The Big Five annotations of the First Impressions V2 dataset are continuous variables. We define personality classes in Section~\ref{sec:personality_classes} because it is a database-specific modification.

We applied MS loss, which is defined as a pair weighting problem, and empirical results show that it is superior over other commonly used loss functions, namely the contrastive loss, triplet loss, binomial deviance loss, and lifted structure loss. To compute a cross-modal MS loss, first, the $e_A$ audio, $e_V$ visual, and $e_T$ textual embeddings are combined to form a triple-sized batch of embeddings denoted as $\{e_A, e_V, e_T\}$, then the similarity metrics are calculated using mixed embeddings of different modalities. 

Similarity is defined between two embeddings $e_1$ and $e_2$ as the dot product of the vectors considering only the $j$th personality trait, denoted as $D_{e_1,e_2}^j=\langle S(e_1), S(e_2) \rangle$.
MS consists of two parts: mining and weighting. Both schemes are integrated into a single loss function, which is defined as follows:
\begin{equation}\label{eq:ms_loss}
\mathcal{L}_{MS} = \frac{1}{5n} \sum_{i=1}^n \sum_{j=1}^5 \Bigg\{ \frac{1}{\alpha}\log\big[1+\sum_{k\in\mathcal{P}_i^j}e^{-\alpha(D_{ik}^j-\lambda)}\big]+  \frac{1}{\beta}\log\big[1+\sum_{k\in\mathcal{N}_i^j}e^{\beta(D_{ik}^j-\lambda)}\big]\Bigg\},
\end{equation}
where $D$ is the similarity matrix within a triple-size mini-batch, $D_{ik}^j$ is the similarity of two embeddings $i$ and $k$, $\mathcal{P}^j$ and $\mathcal{N}^j$ are the sets of positive and negative examples considering only the $j$th trait class labels, respectively. $\alpha$, $\beta$ and $\lambda$ are fixed hyper-parameters.

We calculated a mean, trait-wise multi-similarity loss, considering all 5 target variables per sample within a mini-batch. In the case of non-extreme examples, one or more modalities contain inadequate information to aid the deep embedding process, so we modified the online hard sample mining process to only consider extreme samples as an anchor. In the third learning stage, the $S$ embedding network is trained with trait-wise $\mathcal{L}_{MS}$ with the modified mining procedure. The Siamese network outputs auxiliary vectors that can help the evaluation due to its specific modality mixing mechanism.

\subsection{Fused model}
Our method combines the multimodal regression network and the cross-modal Siamese network in the fourth (and final) learning stage.
The cross-modal embeddings ($e_A, e_V, e_T$) are complementary to the hidden representations of the modality-specific subnetworks s and all of them contribute to the final prediction of $p$.

Model-level fusion is applied, similarly as before: the embeddings are concatenated to the previously fused features, then a $M_2 : \rm I\!R^{3H+3E} \rightarrow \rm I\!R^O$ fully-connected shallow network and $p$ is applied to get the prediction of the Big Five traits. Formally defined as:

\begin{equation}
\hat{y} = p\bigg(M_2 \Big(M_1\big(h_A\oplus h_V\oplus h_T\big)\oplus e_A \oplus e_V \oplus e_T \Big)\bigg),
\end{equation}

\section{Experiments}\label{s:exp_results}

In the following paragraphs, we introduce the dataset used for the experiments, concretize input and hidden dimensions, predetermined hyperparameters during the network implementation. Then the evaluation metric, personality trait class definitions, visualization, and the results are presented.

\subsection{Database}
We used the ChaLearn: First Impressions V2 database for our experiments because it is the largest publicly available in-the-wild dataset in this subfield. 

\begin{figure}[!ht]
\centering
\includegraphics[width=0.85\textwidth]{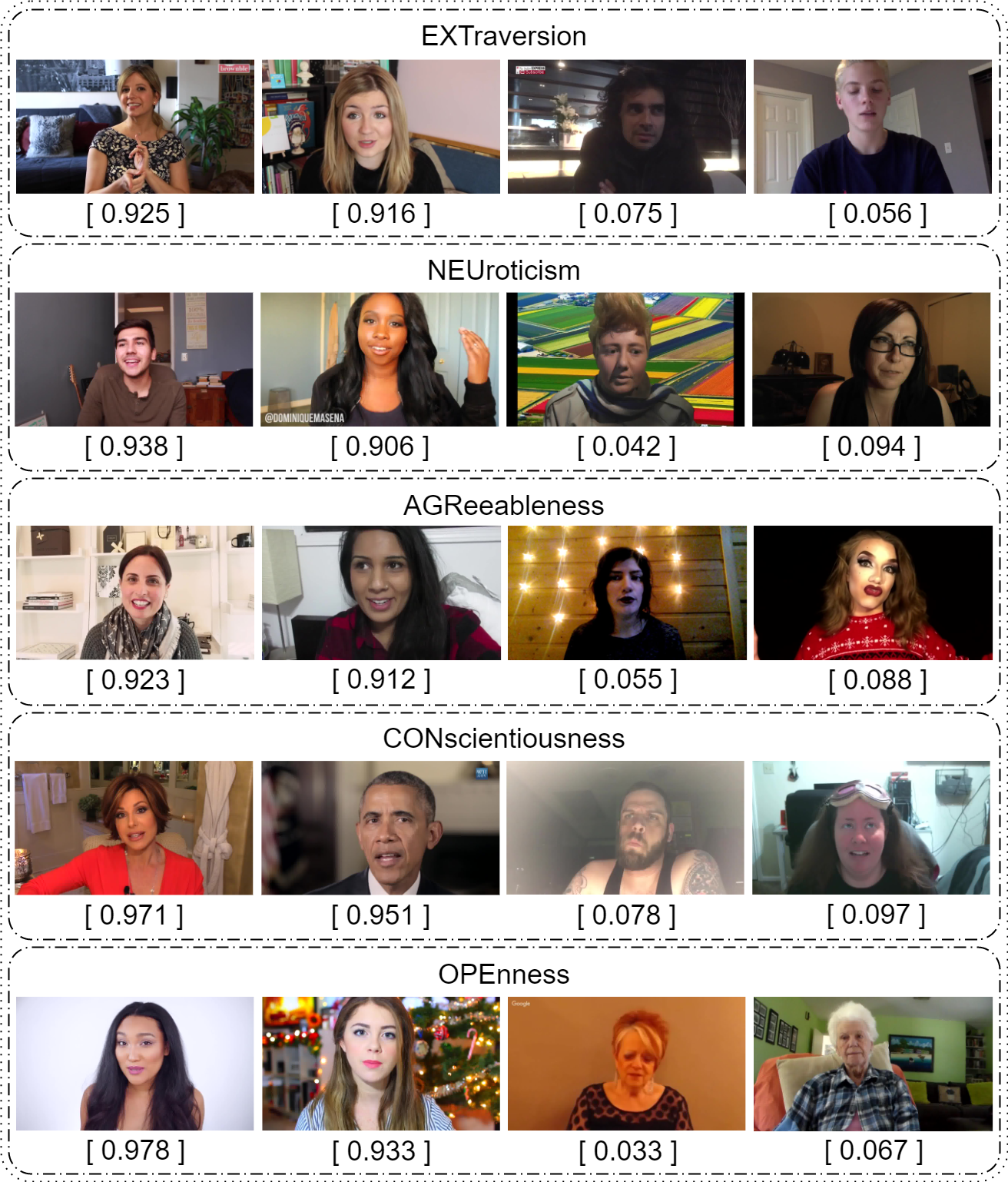}
\caption{Examples of the First Impression V2 dataset. For each video the ground truth Big Five scores are provided.
For each trait, the first two samples instantiate the high extremes, and the last two examples demonstrate the low extremes of a given trait.}\label{fig:samples}
\end{figure}

\newpage

The dataset contains 15 seconds long videos, which are collected automatically. Transcripts of the video clips are generated by a cloud transcription service Rev. The clips are annotated by Amazon Mechanical Turk (AMT) workers using a special interface \cite{db2016}. They registered annotations using pairwise comparisons, and then they converted the votes to cardinal values by fitting a BTL model with maximum likelihood estimation. Values are scaled, so every video sample has five continuous trait scores between 0 and 1. Each trait represents a range bounded by two extremes. For example, for extraversion, the two polar ends are extreme extraversion and extreme introversion, which can be described with the words ``friendly'' and ``reserved'', respectively. A few examples from the dataset focusing on the extreme poles per trait are depicted in Figure~\ref{fig:samples}.

Creators of this dataset rely on the perception of human subjects watching the videos. It is a different task than evaluating real personality traits with experts, but equally useful in the context of human interaction.

One specialty of this dataset is that the target variables have unbalanced data distribution. The regression-to-the-mean problem is emphasized because the scores follow a Gaussian distribution, and the optimization process likely produces predictions near the mean of ground truth values to minimize the loss. We alleviated this problem with the Bell loss \cite{crnet2020}, which is similar to the Mean Squared Error, however, it can produce higher gradients when the prediction is closer to the ground truth.

\subsection{Experimental Setup}\label{s:exp_details}

Our experiments are conducted with Tensorflow on a single GeForce RTX 2080 Ti GPU.

The training process is performed in multiple learning stages. The weights are not modified after a finished stage. We used Adam \cite{adam2014} optimizer with a $0.001$ initial learning rate with a polynomial decay schedule throughout all experiments. Following the work of \cite{crnet2020}, we set the parameters of Bell loss $\sigma=9$ and $\gamma=300$. In the first, second, and fourth learning stages, $\mathcal{L}$ was used as the loss function (Section~\ref{sec:subnet_losses}).

For reduced complexity, we define $Q=256$ and $O=512$ in Section~\ref{s:methodology}. All three modality-specific networks produce 256-dimensional feature vectors, and following a concatenation, shared dense networks produce 512-dimensional vectors in the baseline and the proposed fused model as well.

For acoustic representation 88-dimensional eGeMAPS vectors are used ($N=88$).
We fed a mini-batch of 128 vectors to the audio subnetwork and tuned the two fully-connected layers for 100 epochs with early stopping.

After the frame selection (6 frames per clip) and augmentation techniques \mbox{$6\times128\times128\times3$} input features are fed to the visual subnetwork ($F=6$, $H=W=128$, $C=3$). We trained it from scratch with a mini-batch of 22 video sequences for 80 epochs. Dropout with a $0.5$ rate was applied before the fully-connected layer as an extra regularization.

For semantic word representation, we used 300-dimensional GloVe embeddings. We empirically set the sequence length to 80. After converting every token to its corresponding GloVe vector, an $80\times300$ matrix is produced for every sample ($K=80$, $G=300$).
For the textual subnetwork, we used $0.5$ for the Bi-GRU input dropout rate. We also applied a simplified attention mechanism \cite{attention2015} and tuned the subnetwork for 50 epochs.

The Siamese network consists of two fully-connected hidden layers with 200 neurons each, and a linear dense output layer with 128 units. Dropout with a 0.5 rate was applied after the first hidden layer. In the third learning stage, we used $\mathcal{L}_{MS}$ as the loss function (Equation~\ref{eq:ms_loss}).
We used ReLU as activation function and Kaiming/He normal initialization, in addition to $0.0005$ weight decay in every dense layer, except within the Siamese, where weight decay is not considered. 

\newpage

\subsection{Evaluation metrics}

During the ChaLearn challenge, the $R_{acc}$ (``1-Mean Absolute Error'') was the performance metric, so many publications employed it. It is defined as follows:

\begin{equation}
R_{acc} = 1 - \frac{1}{5n} \sum_{i=1}^{n} \sum_{j=1}^{5} \vert y_{ij}-\hat{y}_{ij} \vert,
\end{equation}

\noindent
where n is the number of samples, $y_{ij}$ and $\hat{y}_{ij}$ are the ground truth and prediction of $i$th sample and $j$th trait.

\subsection{Personality trait class definition}\label{sec:personality_classes}

Annotation regarding the First Impressions V2 dataset consists of 5 continuous variables. Samples can be grouped into different classes by splitting the [0, 1] interval to equal, smaller intervals. In this work, we aim to differentiate extreme examples from ordinary samples based on the ground truth values. We determine 4 classes per trait, and we are focusing on the two extremes, which can be monitored in various clinical sessions later on: the low-extreme and high-extreme classes, which are labeled as C1 and C4, respectively.

\begin{figure}[!ht]
\centering
\includegraphics[width=1.0\textwidth]{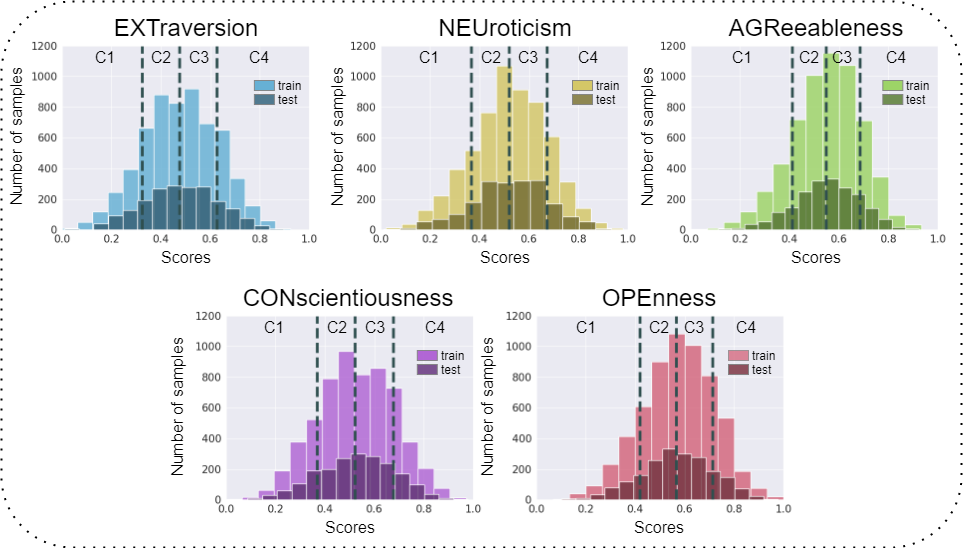}
\caption{Personality trait class definitions. Continuous ground truth values are segmented into 4 classes. The thresholds are determined using the mean and standard deviation calculated on the train set trait-wise. Samples from C1 and C4 are the low extremes and high extremes, respectively.}\label{fig:db_hists}
\end{figure}

However, in our case the ground truth follows a Gaussian distribution, and splitting the [0, 1] interval to equal parts would lead us to an undesirably unbalanced number of extreme samples. To address this issue, we can create more balanced classes by determining the following segmentation thresholds: scores in range $[0, \bar{t}-\sigma_t)$ belong to the low-extreme class (C1), values in range $[\bar{t}-\sigma_t, \bar{t})$ as well as $[\bar{t}, \bar{t}+\sigma_t)$ are labeled as ordinary (C2, C3), and samples between $[\bar{t}+\sigma_t, 1]$ are the high-extremes (C4), where $\bar{t}$ and $\sigma_t$ is the mean and standard deviation calculated over all training samples of $t$ personality trait. Figure \ref{fig:db_hists} demonstrate the class definitions on the histograms of the train and test sets.

\subsection{Visualization of Cross-Modal Embeddings}
\begin{figure}[!ht]
\centering
\includegraphics[width=0.8\textwidth]{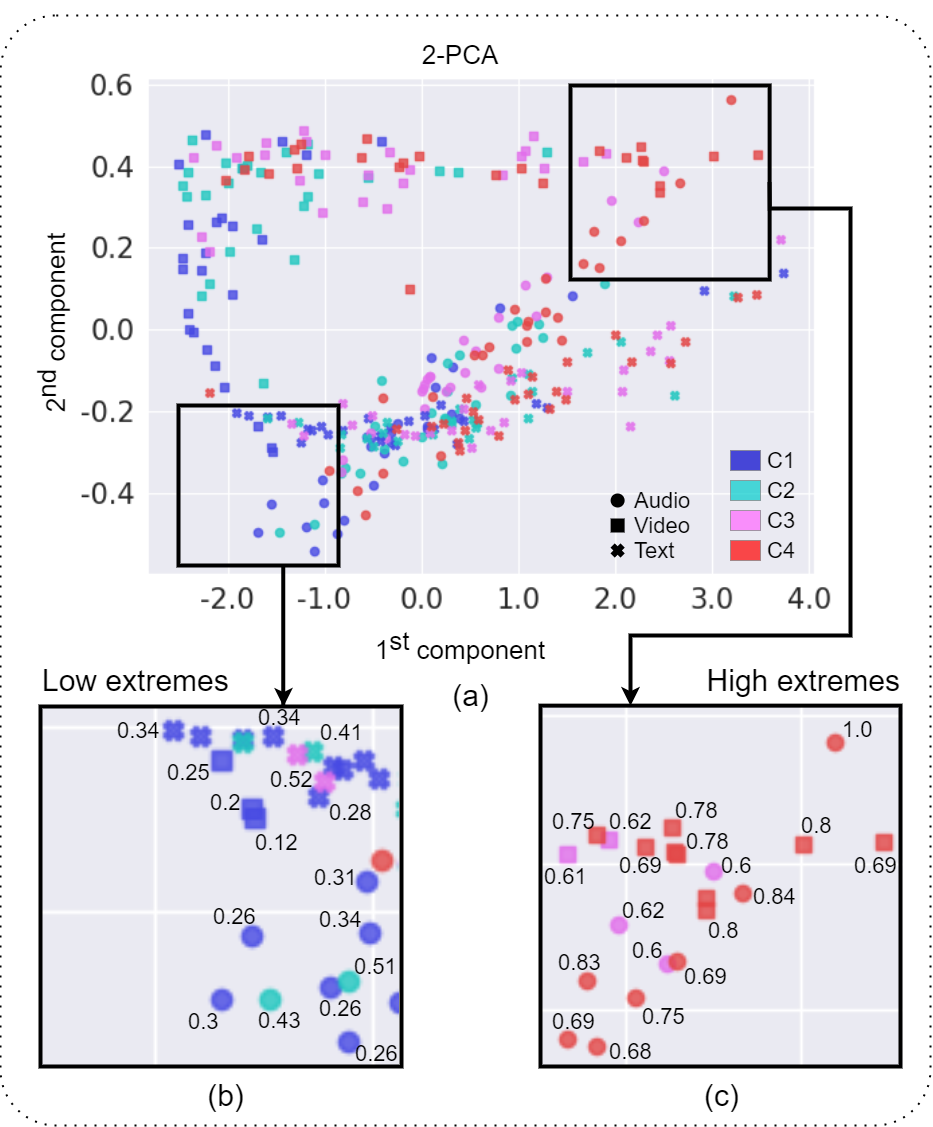}
\caption{Visualization of 2-component PCA of cross- and multimodal embeddings of the ``test'' set (a), showing NEUroticism ground truth values and class labels. The audio, video and text modalities are drawn with circle, square and cross, respectively. The four personality classes are represented with colors, where the blue is the low extreme (C1), and the red is the high extreme class (C4). In the (b) and (c), we emphasize embeddings within the two extreme poles of NEUroticism. }\label{fig:cross_2pca}
\end{figure}

We transformed the acoustic, visual, and textual features to a shared coordinate space with a Siamese network. Figure \ref{fig:cross_2pca} shows a two-component Principal Component Analysis (PCA) calculated on the multimodal inputs as visualization, using only the NEUroticism ground truth values and trait classes within plots.

The test set contains 2000 samples, so considering all three modalities, 6000 embeddings are available. We randomly subsampled to avoid highly overlapped markers and overcrowded visualization, also paying attention to preserve the modality and class balance within the subset: 25 embeddings are selected for every class per modality, so on (a) subplot 300 transformed embeddings are present. The figure shows that even using only two components, the two polar ends of a personality trait are successfully separated. However, there is a continuous transition between trait classes, especially in the case of C2 and C3: the ground truth values are around the mean, and there are hardly perceived or any clues to make these samples more separated using the available inputs.

\section{Results}\label{s:results}

We performed an ablation study with the used modalities to measure the added values of information sources. For the sake of comparison, a prior model obtained directly from the training labels (by averaging) on this dataset was capable of obtaining close to $0.88$ of $R_{acc}$ at test stage~\cite{Escalante2020} due to the highly centralized distribution. In turn, changes in the third digit are relevant. Table \ref{t:ablation} indicates that the video modality contains the most information, with an average score of $0.9074$. Apparent personality traits can be determined accurately using only a single frame: $0.9056$ score over the test set strengthens the statement of trait assignment among human observers can be as fast as 100ms \cite{fastassignment2006}.
The bi-modal systems produce a clear performance jump in every single case compared to the monomodal configurations. Furthermore, the ``Audio + Video + Text'' model performed the expected best result: the different modalities supplement each other. 

Thus, we can fairly compare the proposed method to the ``Audio + Video + Text'' baseline.
Table \ref{t:ablation} shows that our method performs more superior overall, emphasizing the improvement produced by cross-modal embeddings from $0.9094$ to $0.9127$.

\begin{table}[!h]
\centering
\begin{tabular}{l|l|l|l|l|l|l|}
\cline{2-7} 
& \multicolumn{1}{c|}{EXT} & \multicolumn{1}{c|}{NEU} & \multicolumn{1}{c|}{AGR} & \multicolumn{1}{c|}{CON} & \multicolumn{1}{c|}{OPE} & \multicolumn{1}{c|}{Avg} \\ \hline
\multicolumn{1}{|l|}{Audio}              & 0.8947 & 0.8955 & 0.9016 & 0.8916 & 0.9007 & 0.8968 \\
\multicolumn{1}{|l|}{Scene}              & 0.9065 & 0.8990 & 0.9065 & 0.9110 & 0.9048 & 0.9056 \\
\multicolumn{1}{|l|}{Video}              & 0.9086 & 0.9016 & 0.9072 & 0.9132 & 0.9065 & 0.9074 \\
\multicolumn{1}{|l|}{Text}
& 0.8837 & 0.8853 & 0.8982 & 0.8841 & 0.8900 & 0.8882 \\ \hline
\multicolumn{1}{|l|}{Audio + Video} 
& 0.9097 & 0.9041 & 0.9088 & 0.9143 & 0.9074 & 0.9089 \\
\multicolumn{1}{|l|}{Audio + Text}
& 0.8958 & 0.8965 & 0.9023 & 0.8952 & 0.9016 & 0.8983 \\
\multicolumn{1}{|l|}{Text + Video}
& 0.9105 & 0.9041 & 0.9080 & 0.9140 & 0.9073 & 0.9088 \\ \hline
\multicolumn{1}{|l|}{Audio + Video + Text} & 0.9108 & 0.9083 & 0.9108 & 0.9103 & 0.9069 & 0.9094 \\ \hline
\multicolumn{1}{|l|}{$Ours$} & \textbf{0.9142} & \textbf{0.9112} & \textbf{0.9127} & \textbf{0.9154} & \textbf{0.9102} & \textbf{0.9127} \\ \hline
\end{tabular}
\caption{Comparison on the performance $R_{acc}$ of the network trained with different data modalities.}
\label{t:ablation}
\end{table}

We also evaluated the trained system using only extreme samples. We subsampled the test set, so the subset only contained examples from C1 and C4, trait-wise. In Table \ref{t:extreme}, the ``All'' column values are produced on the whole test set by the baseline and our method, respectively. In the case of ``Low'' and ``High'' columns, the corresponding personality trait classes are C1 and C4. The results indicate that we can enhance the prediction of high extreme values at the expense of low extreme prediction in most cases. However, focusing on CONscientiousness, enhanced quality of both low and high extreme predictions can be observed.

\begin{table}[!h]
\centering
\begin{tabular}{|c|ccc|ccc|}
\hline
\multicolumn{1}{|l|}{} & \multicolumn{3}{c|}{Baseline}
& \multicolumn{3}{c|}{Ours}
\\ \hline
& All & Low & High & All & Low & High \\ \hline
EXT & 0.9108 & \textbf{0.8870} & 0.8739 & \textbf{0.9142} & 0.8841 & \textbf{0.8768}     \\ \hline
NEU & 0.9083 & \textbf{0.8730} & 0.8739 & \textbf{0.9112} & 0.8722 & \textbf{0.8742} \\ \hline
AGR & 0.9108 & \textbf{0.8590} & 0.8626 & \textbf{0.9127} & 0.8573 & \textbf{0.8644}     \\ \hline
CON & 0.9103 & 0.8731 & 0.8832 & \textbf{0.9154} & \textbf{0.8753} & \textbf{0.8891} \\ \hline
OPE & 0.9069 & \textbf{0.8691} & 0.8702 & \textbf{0.9102} & 0.8684 & \textbf{0.8794} \\ \hline
\end{tabular}
\caption{Network performance $R_{acc}$ on all samples, low and high extreme examples. Baseline: Audio + Video + Text. Ours: Audio + Video + Text fused with cross-modal embeddings.}
\label{t:extreme}
\end{table}

\section{Conclusions}\label{s:conc}
In this article, we proposed a general-purpose learning framework for the Big Five personality trait prediction, which deals with multimodal data.
We used the currently largest publicly available dataset in our experiments, the ChaLearn: First Impressions V2, and we created modality invariant embeddings to make the different input modalities supplement each other. 

An ablation study has demonstrated the added values of different modalities, as well as the proposed extension. We applied a modified multi-similarity constraint over acoustic, visual, and textual representations to implicitly exploit the mutual information.
Experiments show that we achieved higher overall prediction accuracy, surpassing the performance of baseline multimodal configurations. Besides, we evaluated the proposed method of extreme examples, which produced the desired results in some cases.

To our best knowledge, this is the first work that introduces cross-modal embedding for personality trait prediction. The proposed learning framework is far from perfect. It could be further developed, which is planned for future works. The feature extraction part could be improved to produce more diverse and descriptive representations. Probabilities could be utilized within the triplet constraint to consider the uncertainty around trait class segmentation thresholds properly. The multiple learning phases could be combined to form an end-to-end training process for better useability.

\section*{Acknowledgments}
The research was supported by the Ministry of Innovation and Technology NRDI Office within the framework of the Artificial Intelligence National Laboratory Program.
ÁF and RRS were supported by part through grants EFOP-3.6.3-VEKOP-16-2017-00001 and EFOP-3.6.3-VEKOP-16-2017-00002, respectively. AL was supported by the project “Application Domain Specific Highly Reliable IT Solutions” implemented with the support provided by the National Research, Development and Innovation Fund of Hungary and financed under the Thematic Excellence Programme no. 2020-4.1.1.-TKP2020 (National Challenges Subprogramme) funding scheme.

\bibliographystyle{plain}
\bibliography{references}

\begin{thebibliography}{10}

\bibitem{busso2004analysis}
B.~Carlos, D.~Zhigang, Y.~Serdar, B.~Murtaza, L.~Chulmin, K.~Abe, L.~Sungbok,
  N.~Ulrich, and N.~Shrikanth.
\newblock Analysis of emotion recognition using facial expressions, speech and
  multimodal information.
\newblock In {\em Proceedings of the 6th International Conference on Multimodal
  Interfaces}, pages 205--211, 2014.

\bibitem{digman1990personality}
J.~M. Digman.
\newblock Personality structure: Emergence of the five-factor model.
\newblock {\em Annual Review of Psychology}, 41(1):417--440, 1990.

\bibitem{hugo2018explaining}
H.~J. Escalante, H.~Kaya, A.~Salah, S.~Escalera, Y.~Gucluturk, U.~Guclu,
  X.~Baró, I.~Guyon, J.~Junior, M.~Madadi, S.~Ayache, E.~Viegas,
  F.~Gürpınar, A.~Wicaksana, C.~Liem, M.~Gerven, and R.~Lier.
\newblock Explaining first impressions: Modeling, recognizing, and explaining
  apparent personality from videos.
\newblock {\em IEEE Transactions on Affective Computing}, pages 1--18, 2018.

\bibitem{Escalante2020}
H.~J. Escalante, H.~Kaya, A.~Salah, S.~Escalera, Y.~Güçlütürk, U.~Güçlü,
  X.~Baró, I.~Guyon, J.~C. S.~Jacques Junior, M.~Madadi, S.~Ayache, E.~Viegas,
  F.~Gurpinar, A.~S. Wicaksana, C.~Liem, M.~A. J.~Van Gerven, and R.~Van Lier.
\newblock Modeling, recognizing, and explaining apparent personality from
  videos.
\newblock {\em IEEE Transactions on Affective Computing}, 2020.

\bibitem{eyben2015geneva}
F.~Eyben, K.~Scherer, B.~Schuller, J.~Sundberg, E.~Andr{\'e}, C.~Busso,
  L.~Devillers, J.~Epps, P.~Laukka, S.~Narayanan, and K.~Truong.
\newblock The geneva minimalistic acoustic parameter set (gemaps) for voice
  research and affective computing.
\newblock {\em IEEE Transactions on Affective Computing}, 7(2):190--202, 2015.

\bibitem{eyben2010opensmile}
F.~Eyben, M.~W{\"o}llmer, and B.~Schuller.
\newblock Opensmile: the munich versatile and fast open-source audio feature
  extractor.
\newblock In {\em Proceedings of the 18th ACM international conference on
  Multimedia}, pages 1459--1462, 2010.

\bibitem{bjorn2018emotion}
J.~Han, Z.~Zhang, G.~Keren, and B.~Schuller.
\newblock Emotion recognition in speech with latent discriminative
  representations learning.
\newblock {\em Acta Acustica united with Acustica}, 104(5):737--740, 2018.

\bibitem{han2019emobed}
J.~Han, Z.~Zhang, Z.~Ren, and B.~W. Schuller.
\newblock Emobed: Strengthening monomodal emotion recognition via training with
  crossmodal emotion embeddings.
\newblock {\em IEEE Transactions on Affective Computing}, 2019.

\bibitem{hoffer2015deep}
E.~Hoffer and N.~Ailon.
\newblock Deep metric learning using triplet network.
\newblock In {\em International Workshop on Similarity-Based Pattern
  Recognition}, pages 84--92, 2015.

\bibitem{junior2019first}
J.~C. S.~Jacques Junior, Y.~G{\"u}{\c{c}}l{\"u}t{\"u}rk, M.~P{\'e}rez,
  U.~G{\"u}{\c{c}}l{\"u}, C.~Andujar, X.~Bar{\'o}, H.~J. Escalante, I.~Guyon,
  M.~A. J.~Van Gerven, and R.~Van Lier.
\newblock First impressions: A survey on vision-based apparent personality
  trait analysis.
\newblock {\em IEEE Transactions on Affective Computing}, pages 1--20, 2019.

\bibitem{firstimpression2018}
J.~J. Junior, Y.~G{\"u}çl{\"u}t{\"u}rk, M.~P{\'e}rez, U.~G{\"u}çl{\"u},
  X.~Bar{\'o}, H.~Escalante, I.~Guyon, M.~V. Gerven, R.~Lier, and S.~Escalera.
\newblock First impressions: A survey on vision-based apparent personality
  trait analysis.
\newblock {\em arXiv: Computer Vision and Pattern Recognition}, 2018.

\bibitem{omno2018investigating}
O.~Kampman, E.~J. Barezi, D.~Bertero, and P.~Fung.
\newblock Investigating audio, visual, and text fusion methods for end-to-end
  automatic personality prediction.
\newblock In {\em Proceedings of the 56th Annual Meeting of the Association for
  Computational Linguistics}, pages 606--611, 2018.

\bibitem{kang2015learning}
C.~Kang, S.~Xiang, S.~Liao, C.~Xu, and C.~Pan.
\newblock Learning consistent feature representation for cross-modal multimedia
  retrieval.
\newblock {\em IEEE Transactions on Multimedia}, 17(3):370--381, 2015.

\bibitem{kaya2017multi}
H.~Kaya, F.~Gurpinar, and A.~A. Salah.
\newblock Multimodal score fusion and decision trees for explainable automatic
  job candidate screening from video cvs.
\newblock In {\em Proceedings of the IEEE Conference on Computer Vision and
  Pattern Recognition Workshops}, pages 1--9, 2017.

\bibitem{adam2014}
D.~P. Kingma and J.~Ba.
\newblock Adam: A method for stochastic optimization.
\newblock {\em arXiv preprint}, arXiv:1412.6980, 2014.

\bibitem{crnet2020}
Y.~Li, J.~Wan, Q.~Miao, S.~Escalera, H.~Fang, H.~Chen, X.~Qi, and G.~Guo.
\newblock Cr-net: A deep classification-regression network for multimodal
  apparent personality analysis.
\newblock {\em International Journal of Computer Vision}, pages 1--18, 2020.

\bibitem{andrew2011multimodal}
J.~Ngiam, A.~Khosla, M.~Kim, J.~Nam, H.~Lee, and A.~Y. Ng.
\newblock Multimodal deep learning.
\newblock In {\em Proceedings of the 28th International Conference on Machine
  Learning (ICML)}, pages 689--696, 2011.

\bibitem{db2016}
V.~Ponce-L{\'o}pez, B.~Chen, M.~Oliu, C.~Corneanu, A.~Clap{\'e}s, I.~Guyon,
  X.~Bar{\'o}, H.~J. Escalante, and S.~Escalera.
\newblock Chalearn lap 2016: First round challenge on first impressions -
  dataset and results.
\newblock In {\em Computer Vision – ECCV 2016 Workshops, Lecture Notes in
  Computer Science}, pages 400--418, 2016.

\bibitem{attention2015}
C.~Raffel and D.~P.~W. Ellis.
\newblock Feed-forward networks with attention can solve some long-term memory
  problems.
\newblock {\em arXiv preprint}, arXiv:1512.08756, 2016.

\bibitem{lpmorency2018learning}
Y.-H.~H. Tsai, P.~P. Liang, A.~Zadeh, L.-P. Morency, and R.~Salakhutdinov.
\newblock Learning factorized multimodal representations.
\newblock In {\em International Conference on Learning Representations (ICLR)},
  pages 1--20, 2019.

\bibitem{wang2017adversarial}
B.~Wang, Y.~Yang, X.~Xu, A.~Hanjalic, and H.~T. Shen.
\newblock Adversarial cross-modal retrieval.
\newblock In {\em Proceedings of the 25th ACM international conference on
  Multimedia}, pages 154--162, 2017.

\bibitem{wang2019multi}
X.~Wang, X.~Han, W.~Huang, D.~Dong, and M.~R. Scott.
\newblock Multi-similarity loss with general pair weighting for deep metric
  learning.
\newblock In {\em Proceedings of the IEEE Conference on Computer Vision and
  Pattern Recognition}, pages 5022--5030, 2019.

\bibitem{fastassignment2006}
J.~Willis and A.~Todorov.
\newblock First impressions making up your mind after a 100-ms exposure to a
  face.
\newblock {\em Psychological Science}, 17(7):592--598, 2006.

\bibitem{bjorn2008low}
M.~Wimmer, B.~Schuller, D.~Arsic, G.~Rigoll, and B.~Radig.
\newblock Low-level fusion of audio, video feature for multi-modal emotion
  recognition.
\newblock In {\em Proceedings of the Third International Conference on Computer
  Vision Theory and Applications (VISAPP)}, pages 145--151, 2008.

\bibitem{bjorn2010context}
M.~W{\"o}llmer, A.~Metallinou, F.~Eyben, B.~Schuller, and S.~S. Narayanan.
\newblock Context-sensitive multimodal emotion recognition from speech and
  facial expression using bidirectional lstm modeling.
\newblock In {\em 11th Annual Conference of the International Speech
  Communication Association (INTERSPEECH)}, pages 1--4, 2010.

\bibitem{wang2018}
W.~Xintao, Y.~Ke, D.~Chao, and L.~C. Chen.
\newblock Recovering realistic texture in image super-resolution by deep
  spatial feature transform.
\newblock In {\em IEEE Conference on Computer Vision and Pattern Recognition
  (CVPR)}, 2018.

\bibitem{zadeh2017tensor}
A.~Zadeh, M.~Chen, S.~Poria, E.~Cambria, and L.-P. Morency.
\newblock Tensor fusion network for multimodal sentiment analysis.
\newblock In {\em Proceedings of the 2017 Conference on Empirical Methods in
  Natural Language Processing}, pages 1103--1114, 2017.

\bibitem{lpmorency2017tensor}
A.~Zadeh, M.~Chen, S.~Poria, E.~Cambria, and L.-P. Morency.
\newblock Tensor fusion network for multimodal sentiment analysis.
\newblock In {\em Proceedings of the 2017 Conference on Empirical Methods in
  Natural Language Processing}, pages 1103--1114, 2017.

\bibitem{zeng2008survey}
Z.~Zeng, M.~Pantic, G.~I. Roisman, and T.~S. Huang.
\newblock A survey of affect recognition methods: Audio, visual, and
  spontaneous expressions.
\newblock {\em IEEE transactions on pattern analysis and machine intelligence},
  31(1):39--58, 2008.

\bibitem{zhang2019persemon}
L.~Zhang, S.~Peng, and S.~Winkler.
\newblock \mbox{PersEmoN}: A deep network for joint analysis of apparent
  personality, emotion and their relationship.
\newblock {\em IEEE Transactions on Affective Computing}, pages 1--10, 2019.

\end{thebibliography}

\end{document}